%% file: iclr2025_conference.tex
\documentclass{article} 
\usepackage{iclr2025_conference,times}

\usepackage{amsmath,amsfonts,bm}
\input{math_commands.tex}

\usepackage{hyperref}
\usepackage{url}
\usepackage{ulem}
\usepackage{graphicx}
\usepackage{subcaption}
\usepackage{algorithm}  
\usepackage{algpseudocode}
\usepackage{bbm}
\usepackage{array}
\usepackage{booktabs}
\usepackage{makecell}
\usepackage{enumitem}
\usepackage{graphics} 
\usepackage{authblk}


\title{Learning to Plan Before Answering: Self-Teaching LLMs to Learn Abstract Plans for Problem Solving} 
\author{%
	\textbf{Jin Zhang$^{1,2}$, Flood Sung$^2$, {Zhilin Yang}$^2$, Yang Gao$^1$, Chongjie Zhang$^3$}  \\
	$^1$Institute for Interdisciplinary Information Sciences, Tsinghua University, China \\
	$^2$Moonshot AI \\
	$^3$Washington University in St. Louis \\
	\texttt{jin-zhan20@mails.tsinghua.edu.cn}}

\iclrfinalcopy % Uncomment for camera-ready version, but NOT for submission.
\begin{document}

\maketitle

\begin{abstract}

In the field of large language model (LLM) post-training, the effectiveness of utilizing synthetic data generated by the LLM itself has been well-presented. However, a key question remains unaddressed: what essential information should such self-generated data encapsulate? Existing approaches only produce step-by-step problem solutions, and fail to capture the abstract meta-knowledge necessary for generalization across similar problems. Drawing insights from cognitive science, where humans employ high-level abstraction to simplify complex problems before delving into specifics, we introduce a novel self-training algorithm: LEarning to Plan before Answering (LEPA). LEPA trains the LLM to formulate anticipatory plans, which serve as abstract meta-knowledge for problem-solving, before engaging with the intricacies of problems. This approach not only outlines the solution generation path but also shields the LLM from the distraction of irrelevant details. During data generation, LEPA first crafts an anticipatory plan based on the problem, and then generates a solution that aligns with both the plan and the problem. LEPA refines the plan through self-reflection, aiming to acquire plans that are instrumental in yielding correct solutions. During model optimization, the LLM is trained to predict both the refined plans and the corresponding solutions. By efficiently extracting and utilizing the anticipatory plans, LEPA demonstrates remarkable superiority over conventional algorithms on various challenging natural language reasoning benchmarks.

\end{abstract}
\section{Introduction}
Large Language Models (LLMs) have revolutionized the field of natural language processing, demonstrating remarkable capabilities in handling complex language tasks \citep{achiam2023gpt,zhao2023survey,yang2024qwen2,shahriar2024putting}. While post-training optimization of LLMs demands a substantial volume of data \citep{xiao2023smoothquant,wang2024comprehensive}, recent works reveal that LLMs obtain the potential of generating high-quality synthetic data themselves \citep{zelikman2022star,gulcehre2023reinforced,singh2023beyond,bansal2024smaller}. These works, known as self-training methods, improve the LLM by iterating between generating data with LLMs and optimizing LLMs with the generated data. Self-training methods alleviate the requirement of expensive human annotations and make post-training much more scalable.

A central challenge in self-training is, what essential information should such self-generated synthetic data encapsulate? Despite remarkable progress, this problem has not been well studied. Previous works only generate step-by-step problem solutions, and train the LLM to maximize the log-likelihood of generating these solutions \citep{zelikman2022star,singh2023beyond}. This approach only trains the LLM to memorize knowledge about task-specific solutions, and fails to capture the high-level abstract meta-knowledge necessary for generalization across similar problems. As a consequence, previous self-training methods obtain only limited generalization abilities, and struggle on difficult natural language tasks such as Hendrycks MATH \citep{hendrycks2021measuring}.

\begin{figure}[ht]
\begin{center}
%\framebox[4.0in]{$\;$}
\includegraphics[width=\textwidth]{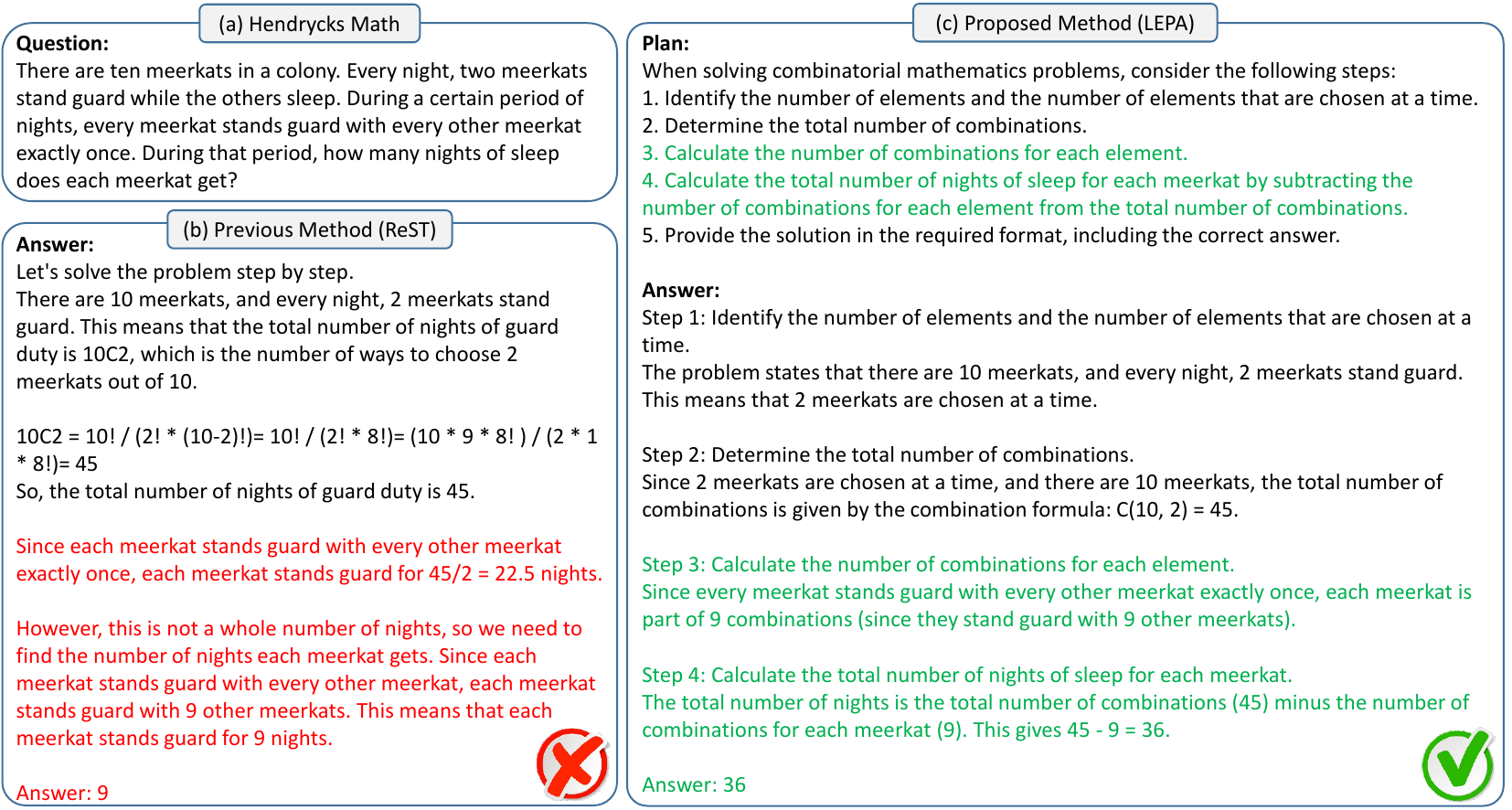}
\end{center}
\caption{A didactic example demonstrating how LEPA outperforms baseline methods by learning to generate anticipatory plans before answering. (a) An example problem in the Hendrycks MATH test set. (b) An incorrect solution given by the LLM trained with a baseline method, ReST. The model fails to generate correct reasoning steps. (c) A correct solution given by the LLM trained with our proposed method, LEPA. The model generates high-quality plans, and then follows the plan to solve the problem correctly.}
\label{fig:lepa-example}
\end{figure}

To tackle this challenge, we draw insights from cognitive science \citep{wang2010cognitive,raduntz2020effect}: humans simplify complex problems through high-level abstraction before engaging with details \citep{ross2009psychology}. Such abstraction not only lightens the cognitive load but also distills high-level meta-knowledge that is transferable to analogous problems. This idea is also evidenced by recent advances in meta-learning \citep{finn2017model,rakelly2019efficient}, which learn generalizable meta-knowledge that enables fast adaptation to similar problems. We propose a novel self-training algorithm, LEarning to Plan before Answering (LEPA), that learns to generate anticipatory plans before generating detailed step-by-step problem solutions. The anticipatory plans serve as high-level abstract meta-knowledge that outlines the solution generation path and shields the LLM from the distraction of irrelevant details.
% LEPA can be interpreted as a meta-learning algorithm that automatically learns to devise these anticipatory plans. 
During data generation, LEPA prompts the LLM to first devise an anticipatory plan that encapsulates the high-level problem-solving steps, and then generate a solution that aligns with both the problem and the plan. If the solution is correct, the plan-solution pair is stored into the training dataset. Otherwise, the LLM is asked to reflect on the plan and the incorrect solution, and refine the plan until it successfully prompts the LLM to generate correct solutions. With this self-reflection mechanism, LEPA acquires plans that are instrumental in yielding
correct solutions. During model optimization, we utilize supervised fine-tuning (SFT) to train the LLM to predict both the plans after self-reflection and the corresponding solutions. As shown in Figure \ref{fig:lepa-example}, after self-training with LEPA, the LLM generates helpful abstract anticipatory plans that outline the solution steps and are generalizable to similar problems, thus achieving better performance than baseline algorithms. LEPA is extensively evaluated on various challenging language reasoning benchmarks including Hendrycks MATH, and significantly outperforms baseline methods.

To summarize, our main contributions are listed as follows:

\begin{enumerate}
    \item We present the fundamental problem of what information should self-generated data encapsulate in the field of LLM self-training.
    \item We propose a novel self-training algorithm, LEPA, that learns to generate anticipatory plans, which serves as high-level abstract meta-knowledge guiding solution generation, before generating detailed problem solutions.
    \item We evaluate LEPA on several challenging language reasoning benchmarks and demonstrate LEPA's superior performance compared to based algorithms.
\end{enumerate}

\section{LEarning to Plan before Answering (LEPA)}

This section introduces LEPA, a novel self-training algorithm that self-trains the LLM to devise high-level anticipatory plans, which serve as abstract solution-generation blueprints, before generating detailed problem solutions. LEPA iterates between a data generation phase and a model optimization phase. In the data generation phase, LEPA generates high-quality plan-solution pairs with self-reflection. In the model optimization phase, LEPA fine-tunes the LLM with the generated data using SFT. Finally, we discuss multiple advantages that the anticipatory plans offer for enhancing the self-training process.

\begin{figure}[t]
% \begin{center}
\centering
\begin{subfigure}[b]{0.95\textwidth}
%\framebox[4.0in]{$\;$}
\includegraphics[width=\textwidth]{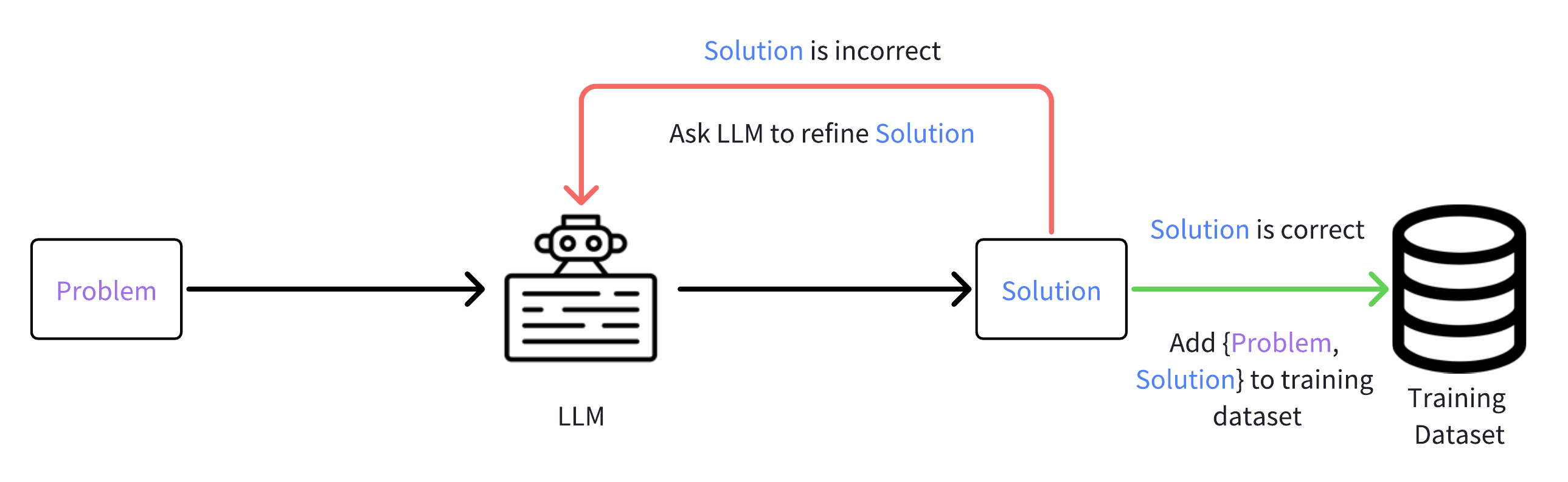}
\caption{Baseline algorithms' data generation procedure.}
\end{subfigure}
\begin{subfigure}[b]{0.95\textwidth}
%\framebox[4.0in]{$\;$}
\includegraphics[width=\textwidth]{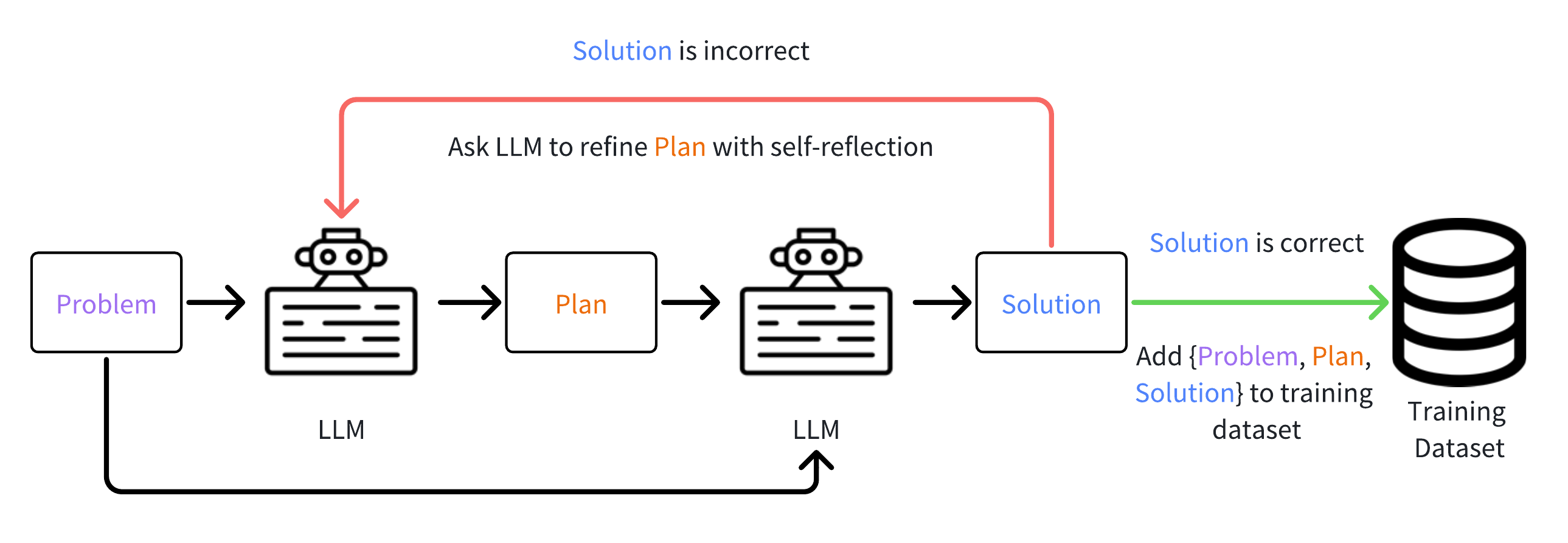}
\caption{LEPA's data generation procedure.}
\end{subfigure}
% \end{center}
\caption{Comparison between baseline algorithms' and LEPA's data generation procedure. (a) Baseline algorithms only generate step-by-step solutions to each problem, lacking high-level abstract meta-knowledge that guides solution generation. (b) LEPA generates anticipatory plans before generating detailed problem solutions. These plans are optimized with self-reflection, and encapsulate the high-level abstract problem-solving steps. The plans efficiently guide the LLM to generate correct solutions.}
\label{fig:alg-generation}
\end{figure}

% \subsection{Preliminaries}

\subsection{Data Generation Phase}
LEPA operates within the common self-training framework, which involves an initial LLM denoted as $\theta_0$, a set of prompts containing $N$ problems $\mathcal{D}_{prompt}=\{x_i\}_{i=0}^{N-1}$, and a binary scoring function $f_{cor}(x_i,y_i)$ that evaluates the correctness of a solution $y_i$ with a score of either 0 or 1.

In each iteration $t$, as depicted in Figure \ref{fig:alg-generation}, LEPA differs from previous methods in that it does not directly prompt the LLM to generate step-by-step solutions to problems. Instead, LEPA instructs the LLM to first generate an anticipatory plan $p_i^t$ that serves as an abstract blueprint for solution generation, and then generate the actual solutions $y_i^t$ based on the plan and the problem. To avoid the degenerate case of generating plans containing detailed step-by-step problem solutions, LEPA stresses in the prompt that the plan should be general high-level meta-knowledge that is applicable to similar problems, and should not contain any problem-specific information such as detailed calculations. If the solution is correct, i.e., $r_{cor}(x_i,y_i)=1$, then the problem-plan-solution tuple $(x_i,p_i^t,y_i^t)$ is added to the training dataset $\mathcal{D}_{train}^{t}$. Otherwise, LEPA refines the plan with self-reflection. The LLM is prompted with the problem, the previous plan, the corresponding incorrect solution, and the correct answer (if accessible). Then LEPA instructs the LLM to reflect on why the previous plan fails to guide itself to generate correct solutions, and then generate a new plan based on its reflection results. To avoid information bypassing, LEPA also stresses in the reflection prompt that the reflected plan should not contain problem-specific information, including detailed calculation and the correct answer. LEPA evaluates the refined plan by instructing the LLM to solve the problem with the refined plan. If the generated solution is correct, the problem-plan-solution tuple $(x_i,p_i^t,y_i^t)$ is added to the training dataset. Otherwise, LEPA repeats the self-reflection process, unless either a correct solution is generated or the number of trials reaches a certain limit $l$. The self-reflection process empowers LLMs to enhance anticipatory plans based on correctness feedback and analysis of unsuccessful attempts, thus efficiently seeking out superior plans.

\subsection{Model Optimization Phase}
In each iteration, after acquiring the training dataset $\mathcal{D}_{train}^{t}$, LPEA optimizes the model with SFT. LEPA formats data into a two-round conversation. In the first round, The user inputs the problem $x_i$ and requires the LLM to generate an anticipatory plan, and the assistant output is the plan $p_i^t$. In the second round, the user instructs the LLM to solve the problem based on the plan it proposed, and the assistant output is the solution $y_i^t$. 
The training objective is to minimize the following negative log-likelihood loss:
\begin{equation}
    \mathcal{L}_{SFT}(\theta_t,\mathcal{D}_{train}^{t})=- \mathbb{E}_{(x_i,p_i^t,y_i^t)\sim\mathcal{D}_{train}^t}[\log p_{\theta_{t}}(p_i^t,y_i^t|x_i)].
\end{equation}

While we employ SFT for algorithm simplicity, LEPA is also compatible with more sophisticated reinforcement learning (RL) algorithms such as Direct Policy Optimization (DPO) \citep{rafailov2024direct} and Proximal Policy Optimization (PPO) \citep{schulman2017proximal}.
We believe RL algorithms can further boost LEPA's performance, and are important future directions. The pseudo-code for LEPA is presented in Algorithm \ref{alg}. Detailed prompts and hyper-parameters used by LEPA is deferred to Appendix \ref{detail}.

\begin{algorithm}
\caption{LEPA: LEarning to Plan before Answering}
\begin{algorithmic}[1] % The number tells where the line numbering should start
\State {\bfseries Require:} An initial LLM $\theta_0$, a set of problems $\mathcal{D}_{prompt}=\{x_i\}_{i=0}^{N-1}$, a binary scoring function $f_{cor}(x_i,y_i)$, number of iterations $T$, maximum self-reflection trials $l$, learning rate $\alpha$
\For{\texttt{$t \gets 0 \textbf{ to } T-1$}} \Comment{In each iteration do}
    \State Initialize an empty training set $\mathcal{D}_{train}^t$ 
    \For{\texttt{$i \gets 0 \textbf{ to } N-1$}} \Comment{For each problem do}
    \State Ask $\theta_t$ to generate anticipatory plan $p_i^{t,0}$ to problem $x_i$
    \State Ask $\theta_t$ to generate solution $y_i^{t,0}$ based on $x_i$ and $p_i^{t,0}$
    \If {$f_{cor}(x_i,y_i^{t,0})$==1} \Comment{Solution is correct, add to training set}
   \State Add $\{x_i,p_i^{t,0},y_i^{t,0}\}$ to $\mathcal{D}_{train}^t$ 
   \Else
   \For{\texttt{$j \gets 1 \textbf{ to } l$}} \Comment{Self-reflection iterations}
   \State Ask $\theta_t$ to self-reflect on $p_i^{t,j-1}$ and $y_i^{t,j-1}$, and generate $p_i^{t,j}$
   \State Ask $\theta_t$ to generate solution $y_i^{t,j}$ based on $x_i$ and $p_i^{t,j}$
   \If {$f_{cor}(x_i,y_i^{t,j})$==1}
   \State Add $\{x_i,p_i^{t,j},y_i^{t,j}\}$ to $\mathcal{D}_{train}^t$ 
   \State \textbf{Break} \Comment{Solution is correct, stop self-reflection}
   \EndIf
   \EndFor
   \EndIf
    \EndFor
    \State $\theta_{t+1} \gets \theta_{t}-\alpha \nabla_{\theta_t}\mathcal{L}_{SFT}(\theta_t,\mathcal{D}_{train}^{t})$ \Comment{Model Optimization with SFT}
\EndFor
\end{algorithmic}
\label{alg}
\end{algorithm}

% align with human reduce cognitive workload
% generalizable high-level knowledge
\subsection{Why is the Anticipatory Plan Beneficial?}
\label{why}
Central to LEPA's efficacy is the anticipatory plan, offering multiple advantages for self-training. This subsection discusses these benefits in detail.

\paragraph{Reducing cognitive workload.} As demonstrated in Figure \ref{fig:lepa-example}, without the anticipatory plans, the LLM may get lost in the problem-solving process, leading to erroneous solution steps. In contrast, the anticipatory plans serve as blueprints that outline the necessary problem-solving steps, and shield the LLM from the distraction of irrelevant details. Consequently, when generating detailed problem solutions, the LLM is conscious of what to do at the current step, and successfully solves the problem. Research in cognitive science \citep{wang2010cognitive,raduntz2020effect} supports the notion that such a structured approach significantly eases cognitive load and improves learning efficiency.

\paragraph{Learning generalizable high-level meta-knowledge.} The anticipatory plans are abstract high-level meta-knowledge that does not involve problem specifics, and is thus generalizable across similar problems. For example, the plan demonstrated in Figure \ref{fig:lepa-example} can be readily adapted to a variety of combinatorial mathematical problems with similar underlying structures but different parameters. From the meta-learning perspective, LEPA can be interpreted as a meta-learning algorithm that extracts the meta-knowledge in the form of anticipatory plans. The learned meta-knowledge empowers the LLM to solve similar problems more effectively.

{\paragraph{Preventing information bypassing.}} When the correct answer is accessible, the anticipatory plans enable self-reflection that avoids the pitfall of information bypassing. Previous methods like STaR \citep{zelikman2022star} directly modify incorrect solutions by referring to the correct answer, and are very likely to {cheat by only modifying the final answer and ignoring the consistency between intermediate steps and the final answer \citep{singh2023beyond}.}  In contrast, as LEPA requires the anticipatory plans to not include any problem-specific information including the final correct answer, it isolates the correct answer from solution generation. {The model must generate correct solutions without seeing the correct answer}, preventing the model from cheating during solution generation. 

% \textbf{Efficient utilization of inference compute.} As proposed by \citet{snell2024scaling} and confirmed by the powerful GPT O1 model \citep{hu2024can}, scaling inference compute can substantially boost LLM performance. LEPA trains the LLM to generate anticipatory plans before generating detailed problem solutions, which is an efficient way to utilize inference compute. We validate this point with empirical experiments results in  
% plans capture high-level general knowledge about problem-solving, enabling the LLM to summarize abstract knowledge that are helpful for solving similar tasks. Secondly,

% Finally, LEPA trains the LLM to output both the plan and the solution in deployment, which efficiently utilizes inference compute to boost performance, which correspond to the inference scaling law proposed by \citet{snell2024scaling} and confirmed by the GPT O1 model \citep{hu2024can}.

% \subsection{Didactic Example}
% This subsection demonstrates a didactic example of how LEPA utilizes self-reflection to optimize the plan in the data generation phase.

\section{Experiments}
To demonstrate the effectiveness of LEPA, we evaluate on several challenging reasoning benchmarks, including Hendrycks MATH (challenging math problems) \citep{hendrycks2021measuring}, Hellaswag (sentence completion reasoning) \citep{zellers2019hellaswag}, BoolQ (paragraph understanding and reasoning) \citep{clark2019boolq}, and PIQA (physics reasoning) \citep{bisk2020piqa}. For Hendrycks MATH, we evaluate solution correctness with the function provided by the dataset creators (\url{https://github.com/hendrycks/math}).We utilize Llama 3 8B Instruct \citep{dubey2024llama} as the initial LLM. LEPA is compared against several representative self-training algorithms: $ReST$ \citep{gulcehre2023reinforced}, $ReST^{EM}$ \citep{singh2023beyond}, and STaR \citep{zelikman2022star}. All these baseline methods only generate step-by-step solutions to problems. Both $ReST$ and $ReST^{EM}$ generate solutions with rejection sampling. In each iteration, $ReST$ fine-tunes the model trained after the previous iteration, while $ReST^{EM}$ instead fine-tunes from the initial LLM. STaR generates solutions by prompting the LLM to modify incorrect solutions with the aid of correct answers, and also fine-tunes from the initial LLM in each iteration. We demonstrate algorithms' test accuracy at convergence\footnote{As STaR's test accuracy drops significantly on MATH, we instead demonstrate its highest test accuracy.}. For a fair comparison, all methods do not utilize few-shot examples in their prompts. We also demonstrate the initial LLM's efficacy, with either a zero-shot CoT prompt \citep{kojima2022large} or a LEPA prompt that instructs it to first generate an anticipatory plan before answering.
\begin{table}[t]
% \small
% \vskip -0.1in
\centering
\caption{Test accuracy of LEPA and various baselines on four challenging reasoning benchmarks. ``CoT'' and ``Plan+CoT'' refer to the initial LLM's performance with a zero-shot CoT prompt and the LEPA prompt, respectively. LEPA demonstrates superior accuracy in comparison to all other algorithms on each of the benchmarks. Numbers in the parentheses are LEPA's performance improvement over the best-performing baseline algorithm on each benchmark.}
% \vspace{\baselineskip}
% \small
\begin{tabular}{c|cc|ccc|c}
\toprule
& \makecell[c]{CoT} & \makecell[c]{Plan+CoT}& $ReST$ & $ReST^{EM}$ & STaR & LEPA \\ \midrule
Hellaswag&60.8\% & 56.1\%&86.3\% & 86.4\% & 85.7\% & \textbf{91.2\% (+4.8\%)}  \\ 
Hendrycks MATH&19.5\% & 22.1\%& 28.2\% & 27.2\% & 25.9\%  & \textbf{30.2\% (+2.0\%)} \\
BoolQ&77.3\% & 80.8\%&84.5\% & 86.3\% & 85.8\% & \textbf{88.4\% (+2.1\%)}  \\
PIQA&67.0\% & 75.7\%&81.4\% & 83.5\% & 84.2\% & \textbf{85.9\% (+1.7\%)}  \\
\midrule
Average&56.1\%&58.7\%&70.1\%&70.8\%&70.4\%&\textbf{73.9\% (+3.1\%)} \\
% Average over 50 tasks& \textbf{0.65$\pm$0.07}&0.33$\pm$0.02&\uline{0.39$\pm$0.03} &0.27$\pm$0.03 &\uline{0.40$\pm$0.04}\\
\bottomrule
\end{tabular}
\label{mainresults}
\end{table}

\subsection{Main Results}
\label{31}
Table \ref{mainresults} presents a comparative analysis of algorithm performance across the four reasoning benchmarks. Notably, in the absence of self-training, the LEPA prompt (Plan+CoT) enhances the initial LLM's performance on three benchmarks when compared to the traditional zero-shot CoT prompt (CoT). This suggests that the practice of formulating anticipatory plans before generating detailed solutions can significantly improve model efficacy. However, on the Hellaswag benchmark, Plan+CoT falls short of CoT, implying that such enhancement is not uniformly achievable across different tasks, potentially due to the initial LLM's lack of calibration for producing high-quality anticipatory plans.
As for self-training performance, baseline self-training algorithms only train the LLM to predict step-by-step solutions, lacking abstract high-level meta-knowledge about problem-solving. As a consequence, these algorithms perform poorly on these benchmarks. In contrast, LEPA efficiently extracts high-level abstract meta-knowledge with the anticipatory plans, thereby surpassing all baseline algorithms consistently across all benchmarks.

\begin{figure}[t]
    \centering
    \begin{subfigure}[b]{0.45\textwidth}
        \includegraphics[width=\textwidth]{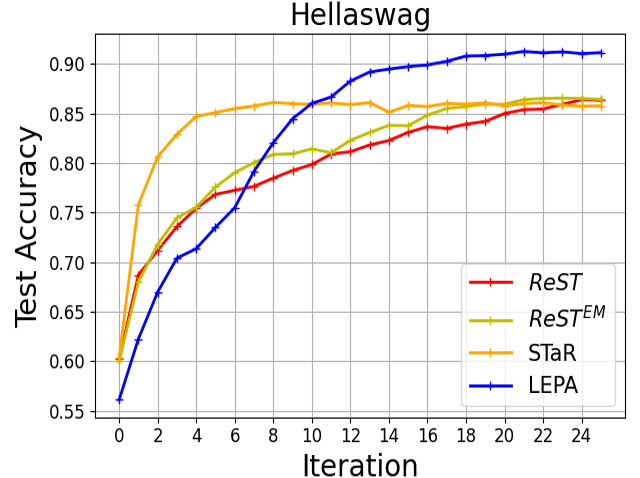}
        % \caption{1}
        \label{fig:subfig1}
    \end{subfigure}
    % \hfill % 可选，用于添加一些水平空间
    \begin{subfigure}[b]{0.45\textwidth}
        \includegraphics[width=\textwidth]{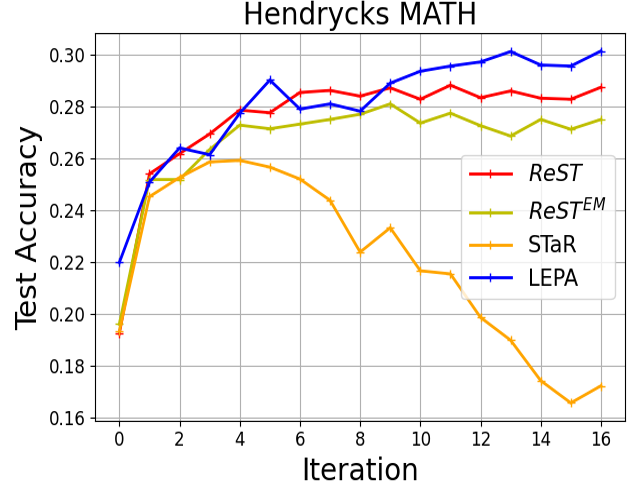}
        % \caption{2}
        \label{fig:subfig2}
    \end{subfigure}
    \begin{subfigure}[b]{0.45\textwidth}
        \includegraphics[width=\textwidth]{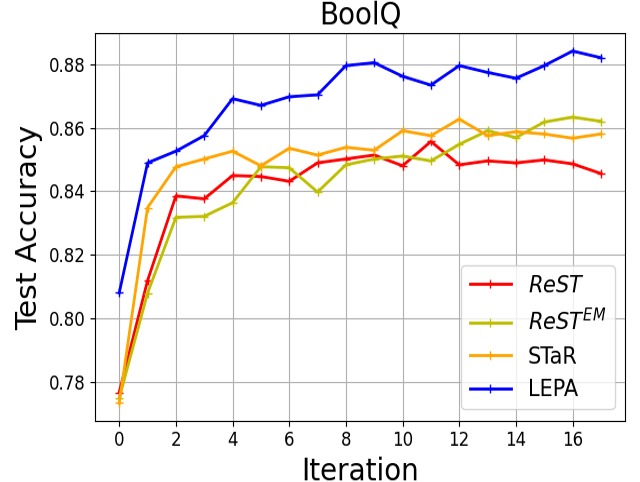}
        % \caption{1}
        \label{fig:subfig3}
    \end{subfigure}
    % \hfill % 可选，用于添加一些水平空间
    \begin{subfigure}[b]{0.45\textwidth}
        \includegraphics[width=\textwidth]{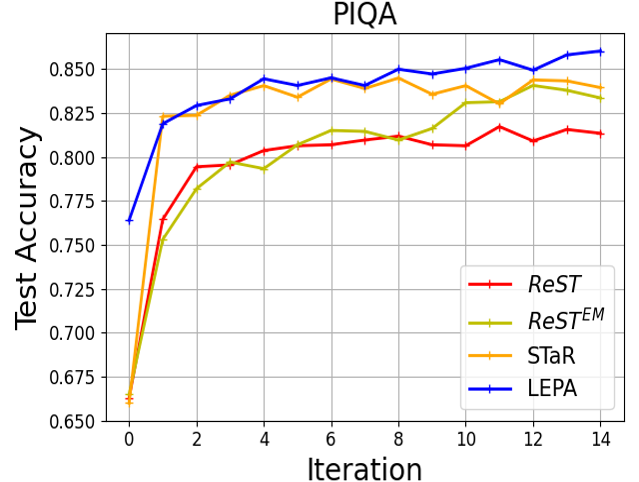}
        % \caption{2}
        \label{fig:subfig4}
    \end{subfigure}
    \vskip -0.1in
    \caption{Algorithms' learning curves on the four benchmarks. LEPA achieves better performance than baseline algorithms.}
    \label{fig:test}
\end{figure}

Figure \ref{fig:test} illustrates algorithms' learning curve across learning iterations. LEPA's superior performance is evident across all benchmarks. Specifically, on Hellaswag, LEPA lags initially during the early iterations (0-10), where the LEPA prompt is slightly less effective than the zero-shot CoT prompt. However, as training progresses, LEPA's performance incrementally surpasses that of the baseline algorithms, suggesting that self-training is instrumental in awakening the LLM's capacity to conceive and leverage anticipatory plans effectively.
On the remaining three benchmarks, LEPA acquires better initial performance and converges at higher test accuracies, demonstrating the effectiveness of introducing the anticipatory plans. We also observe a great performance drop of STaR on Hendrycks MATH. This is because STaR is very likely to generate false-positive solutions, i.e., solutions with wrong rationales but correct final answers \citep{singh2023beyond}, and greatly hinders learning on complex reasoning benchmarks like Hendrycks MATH.

% \begin{figure}[ht]
% \begin{center}
% %\framebox[4.0in]{$\;$}
% \includegraphics[width=0.5\textwidth]{figures/BoolQ.png}
% \end{center}
% \caption{BoolQ learning curve.}
% \label{fig:boolq}
% \end{figure}

% \subsection{Train-Test Gap}

\subsection{Ablation Studies}
\label{32}
LEPA consists of three key components: the anticipatory plan, plan optimization with self-reflection, and utilizing more inference compute to achieve better performance. This subsection discusses the necessity of each component with ablation studies.

\paragraph{Anticipatory plans.} We test a variant of LEPA that does not introduce anticipatory plans in the data generation phase, and only trains the LLM to predict the step-by-step solutions optimized with self-reflection. As shown in Table \ref{abl}, this variant (``Without Plan'') under-performs LEPA. There are two reasons for this degrade in performance. Firstly, without the anticipatory plans, the LLM does not learn abstract high-level meta-knowledge about problem-solving. Secondly, as discussed in Section \ref{why}, directly performing self-reflection on the solutions is very likely to generate false-positive solutions, which greatly hiders learning.

\begin{table}[t]
% \small
\centering
\caption{Ablation study on the anticipatory plan and self-reflection. We also demonstrate the performance of $ReST^{EM}$, the baseline with the highest average test accuracy. ``Without Plan'' is LEPA without anticipatory plans, and ``Without Self-Reflection'' is LEPA without self-reflection.}
% \vspace{\baselineskip}
\begin{tabular}{c|c|c|c|c}
\toprule
% \small
 & $ReST^{EM}$ & LEPA & Without Plan & Without Self-Reflection \\ \midrule
Hendrycks MATH& 27.2\% & \textbf{30.2\%} & 24.3\% & 28.8\% \\
BoolQ & 86.3\%& \textbf{88.4\%} & 84.8\% & 86.9\%  \\
PIQA & 84.2\% & \textbf{85.9\%} & 84.5\% &  84.8\% \\
% \midrule
% Average over 50 tasks& \textbf{0.65$\pm$0.07}&0.33$\pm$0.02&\uline{0.39$\pm$0.03} &0.27$\pm$0.03 &\uline{0.40$\pm$0.04}\\
\bottomrule
\end{tabular}
\label{abl}
\end{table}

\begin{table}[t]
% \small
\centering
\caption{Ablation study on ways of utilizing inference compute. We test on the Hendrycks MATH dataset.``Silence token'' is the variant that adds silence tokens in the solution. ``Correction'' is the variant that trains the LLM to output new solutions if it finds its initial solution incorrect. ``Long Solution'' is the variant that instructs the LLM to generate long solutions. ``\# of Tokens'' is the average token length of the LLM's responses to test problems, and ``Accuracy'' is the LLM's test accuracy. LEPA is the only method that efficiently utilizes additional inference compute to outperform baseline methods. We put the results in two rows due to the page width limit.}
% \vspace{\baselineskip}
\begin{tabular}{cc|cc|cc}
\toprule
\multicolumn{2}{c}{STaR} & \multicolumn{2}{|c|}{$ReST$} & \multicolumn{2}{c}{LEPA} \\
\midrule
\# of Tokens & Accuracy & \# of Tokens & Accuracy & \# of Tokens & Accuracy\\ \midrule
 175.1 & 25.9\% & 477.8 & 28.2\% &  826.4 & \textbf{30.2\%}  \\
\midrule
\multicolumn{2}{c}{Silence Tokens} & \multicolumn{2}{|c|}{Correction} & \multicolumn{2}{c}{Long Solution}  \\
\midrule
% \small
 \# of Tokens & Accuracy & \# of Tokens & Accuracy & \# of Tokens & Accuracy\\ \midrule
869.3 & {28.3\%}   & 979.4 & {27.8\%} & 1409.7 & 25.4\% \\
% \midrule
% Average over 50 tasks& \textbf{0.65$\pm$0.07}&0.33$\pm$0.02&\uline{0.39$\pm$0.03} &0.27$\pm$0.03 &\uline{0.40$\pm$0.04}\\
\bottomrule
\end{tabular}
\label{abl:token}
\end{table}

% \begin{tabular}{c|c|c|c|c|c}
% \toprule
% {STaR} & {$ReST$} & {LEPA} & {Silence Tokens} & {Correction} & {Long Solution}  \\
% \midrule
% \# of Tokens & \# of Tokens &\# of Tokens &\# of Tokens &\# of Tokens &\# of Tokens \\
% \midrule
%  175.1  & 477.8 &  826.4 & 869.3  & 979.4 & 1409.7 \\
% \midrule
% Accuracy & Accuracy & Accuracy & Accuracy & Accuracy & Accuracy  \\
% \midrule
% % \small
% 25.0\% & 28.2\% & \textbf{30.2\%} & {28.3\%}   & {27.8\%}  & 25.4\% \\
% % \midrule
% % Average over 50 tasks& \textbf{0.65$\pm$0.07}&0.33$\pm$0.02&\uline{0.39$\pm$0.03} &0.27$\pm$0.03 &\uline{0.40$\pm$0.04}\\
% \bottomrule
% \end{tabular}
% \label{abl:token}
% \end{table}

\paragraph{Self-reflection.} To demonstrate the necessity of self-reflection in LEPA's plan optimization, we test a variant that instead utilizes rejection sampling \citep{singh2023beyond} to sample plan-answer pairs. As shown in Table \ref{abl}, this variant (``Without Self-Reflection'') also performs worse than LEPA. This result implies that self-reflection is more effective than rejection sampling in optimizing the anticipatory plans, as it gives linguistic feedback for LLMs to improve the previous plans.   %todo

\paragraph{Different ways of utilizing inference compute.} LEPA generates both anticipatory plans and problem solutions, utilizing more compute at inference time. it is worth discussing how much contribution the extra compute makes, and whether the anticipatory plan is an effective way to utilize inference compute. For the first question, as discussed in Section \ref{32}, without self-training, utilizing inference compute with anticipatory plans can improve performance on three of the four benchmarks, and degrade performance on one benchmark. In contrast, after self-training, the anticipatory plans can consistently help LEPA outperform baseline methods. This result demonstrates that extra inference compute contributes a part to LEPA's performance, and self-training is also vital for unlocking the LLM's ability to efficiently utilize these extra compute. For the second question, we test three other variants that train the LLM to utilize inference compute in different ways. The first variant adds silence tokens in the solution to give the LLM more compute to generate answers \citep{goyal2023think}. The second variant trains the LLM to first output a solution, and then outputs a new solution if it finds the original solution incorrect. For data generation of this variant, solutions are generated with rejection sampling, analogous to $ReST$. We synthesize training data by appending correct solutions to the end of incorrect solutions. The third variant simply asks the LLM to generate long solutions. All variants fine-tune the LLM with $ReST$. As shown in Table \ref{abl:token}, LEPA is the only method that successfully utilizes additional inference compute to outperform baseline methods. In contrast, the first variant performs similarly to the $ReST$ baseline, suggesting that silence tokens offer limited benefits for the challenging Hendrycks MATH benchmark. Both the second and the third variant underperform $ReST$, as the LLM is trained to predict tokens with scant correlation to correct solution formulation. The results above implies that introducing the anticipatory plans is a more efficient way to generate long self-training data compared to the aforementioned alternatives. Detailed implementation of these variants are deferred to Appendix \ref{variant}.
% and proper structure needs to be designed.
% outperforms both variants on Hendrycks MATH, which implies that the anticipatory plan is indeed an efficient way to utilize inference compute.

\begin{figure}[t]
\begin{center}
%\framebox[4.0in]{$\;$}
\includegraphics[width=\textwidth]{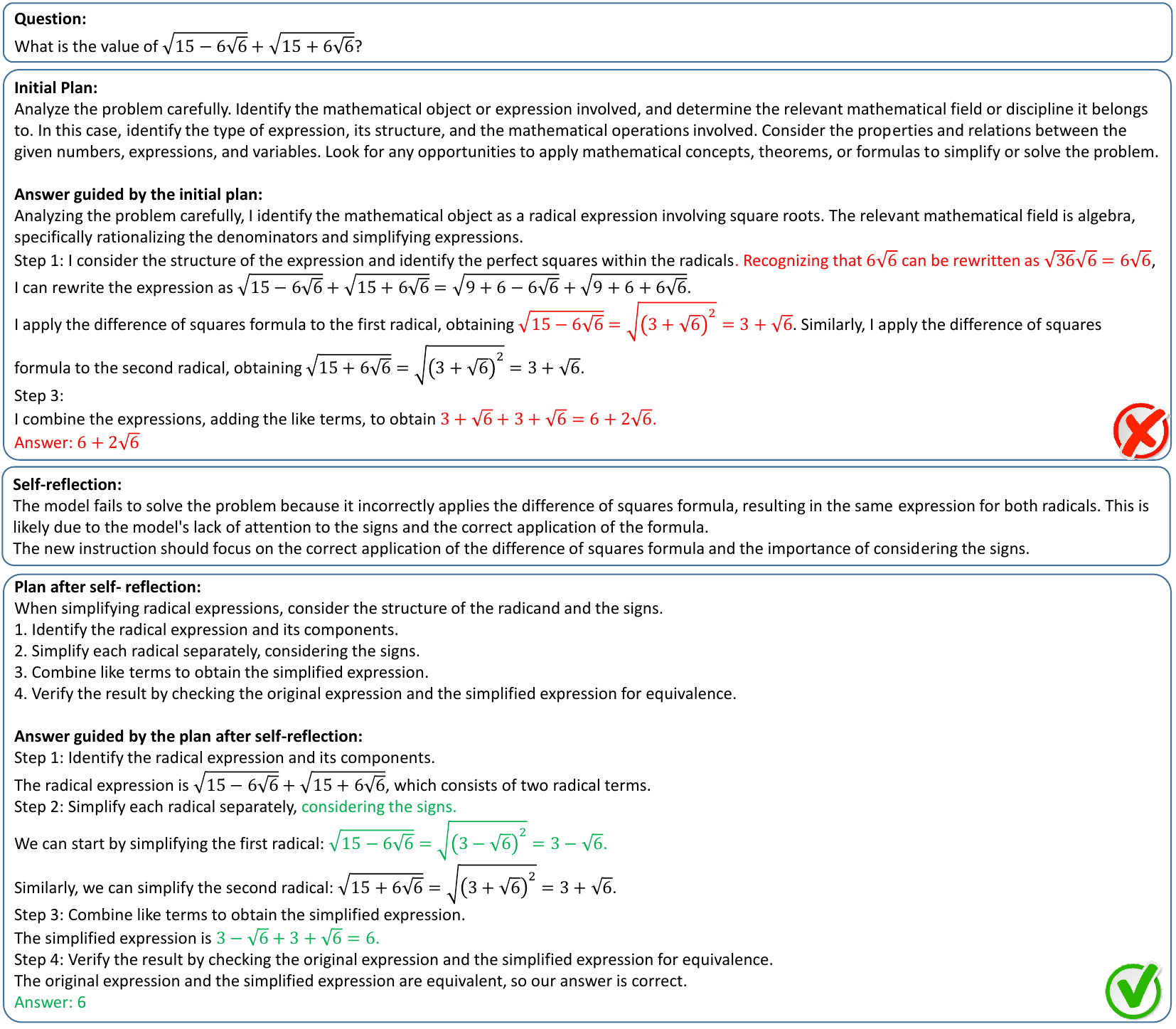}
\end{center}
\caption{A case study demonstrating how LEPA optimizes the anticipatory plans and the solutions with self-reflection. The initial plan is too broad and lacks detail, and fails to provide enough guidance to generate correct answers. The self-reflection process successfully analyses what is wrong, and generates a new, high-quality plan that provides more guidance while maintaining generalizability. With the new plan after self-reflection, the model successfully generates correct solutions.}
\label{fig:case}
\end{figure}

\paragraph{Incorporation with RL algorithms.} To demonstrate that LEPA is also applicable to more advanced RL optimization algorithms, we test a variant of LEPA that utilizes REINFORCE \citep{zhang2021sample} as the underlying optimization algorithm, which is called LEPA+REINFORCE. The only difference between LEPA and LEPA+REINFORCE is that LEPA+REINFORCE labels data with rewards of either 1 or -1 (based on the final answer correctness), and optimizes the LLM with the labelled data using the REINFORCE algorithm. On Hendrycks MATH, LEPA+REINFORCE achieves a test accuracy of 30.6\%, while the original LEPA achieves 30.2\%. This performance improvement demonstrates the potential of incorporating LEPA with more advanced optimization algorithms like RL, and is a promising future direction.

\paragraph{LPEA with RL.} To investigate the potential of incorporating LPEA with RL algorithms, we demonstrate a variant of LEPA that utilizes REINFORCE as the optimization algorithm in the model optimization phase. This variant is named LEPA+REINFORCE. The only difference from LEPA is that LEPA+REINFORCE does not discard failure data. Instead, it utilizes the solution correctness as the reward (1 for correct solutions, -1 for incorrect solutions). This implementation makes no modification to the data generation process. On Hendrycks MATH, LEPA+REINFORCE achieves a test accuracy of 30.6\%, while LEPA achieves 30.2\%. This performance improvement demonstrates the feasibility and effectiveness of incorporating LEPA with RL algorithms. Additional ablation studies including algorithm performance on OOD benchmarks, other LLMs, additional benchmarks, and evaluation with Simple-Eval are deferred to Appendix \ref{aas}.

\subsection{Case Study}
We present a case to demonstrate how LEPA's self-reflection mechanism optimizes the anticipatory plans and the solutions. As shown in Figure \ref{fig:case}, the initial plan generated by the model is too vague, and cannot provide enough guidance for the model to solve the problem correctly. Consequently, during solution generation, the model generates irrelevant steps, makes a mistake in the symbol of the expression, and fails to answer correctly. In the self-reflection process, the model finds out that the previous answer failed to calculate the correct symbols. So it modifies the plan to contain more detailed instructions on how to solve this problem. Note that the plan after self-reflection is still general meta-knowledge that is applicable to a wide range of similar problems. With this modified plan, the model pays more attention to signs, generates only necessary steps, and successfully generates a correct solution. 

\section{Related Works}
\textbf{Self-training.} With the fast development of LLMs, the thirst for data continues to grow. A promising way is to generate high-quality data with the LLM itself. A branch of works mainly focus on designing the data generation progress. STaR \citep{zelikman2022star} operates by initially prompting the LLM to produce step-by-step solutions, followed by an adjustment phase where the LLM corrects its errors with the aid of the correct answers. One severe limitation of STaR is that the modification process makes it very possible to generate false-positive solutions, i.e., solutions with wrong rationales but correct final answers.
RFT \citep{yuan2023scaling}, $ReST$ \citep{gulcehre2023reinforced}, and $ReST^{EM}$ \citep{singh2023beyond} instead adopt rejection sampling for data generation, and suffer less from the false-positive issue.
TRICE \citep{hoffman2024training} improves over STaR by utilizing a Markov-chain Monte Carlo expectation-maximization algorithm to sample solutions, and introducing a control-variate method to control gradient variance. Re-ReST \citep{dou2024reflection} utilizes self-reflection to correct the generated wrong answers. LMSI \citep{huang2022large} considers the scenario where the correctness of model-generated data cannot be verified during training, and filers data with majority voting. 
% Another branch of works explore utilizing self-generated data to train both a generator and a verifier.
% V-STaR \citep{hosseini2024v} utilizes the verifier to select solutions at inference time, while ReST-MCTS \citep{zhang2024rest} utilizes the verifier as a reward function to guide MCTS search.
Apart from these methods,
SPAG \citep{cheng2024self} generates data by asking LLMs to self-play in adversarial games.
These previous methods above only generate step-by-step solutions to problems, and lack high-level meta-knowledge that are generalizable across similar problems. In contrast, LEPA learns abstract meta-knowledge in the form of anticipatory plans, and achieves better performance on complex benchmarks. % todo rl methods

\textbf{Scaling inference compute.} As
proposed by \citet{snell2024scaling} and confirmed by the recent inspiring GPT O1 model \citep{hu2024can}, scaling inference compute can further boost LLM performance. Similar to LEPA, PS Prompting \citep{wang2023plan} also scales inference compute by asking the LLM to first generate a plan before answering, but does not consider how to generate data and fine-tune the LLM. Moreover, it does not consider how to automatically optimize the anticipatory plans. HSP \citep{fu2024hint} is the most relevant work to ours, which trains the LLM to output hints before solving the problem. However, HSP's hints are pre-collected rather than self-generated, and induce additional data collection costs. PHP \citep{zheng2023progressive} utilizes previously generated answers as hints, and encourages the LLM to answer with reference to its previous answers. LEPA efficiently utilizes inference compute by training the LLM to generate helpful anticipatory plans, which contain high-level meta-knowledge on problem-solving, before generating actual problem solutions. These plans are automatically optimized by the LLM itself, and do not require additional human design.

\textbf{Meta-learning.} Meta-learning aims at ``learning to learn'', i.e., designing meta-algorithms that optimize learning algorithms automatically \citep{finn2017model,sung2017learning,rakelly2019efficient,zhang2021metacure,wang2023offline}. LEPA can be interpreted as a meta-learning algorithm that learns the meta-knowledge of designing the anticipatory plans for each problem, rather than designing plans with human effort. The most relevant work is Quiet-STaR \citep{zelikman2024quiet}, which meta-learns meta-tokens that help the LLM to predict the next token. LEPA considers the setting of problem-solving rather than general next-token prediction, and meta-learns the generation of anticipatory problem-solving plans.

\textbf{Planning in LLMs.} Recently, several works have demonstrated the effectiveness of integrating planning in LLMs. ReAct \citep{yao2022react} and DEPS \citep{wang2024describe} generate plans before dealing with decision-making problems, and LUMOS \citep{yin2023lumos} fine-tunes the LLM on pre-collected datasets containing planning data. To our best knowledge, LEPA is the first work to integrate planning in the process of self-training, and improves the LLM's planning ability by training on self-generated data.

\textbf{Self-reflection.} Self-reflection enables LLMs to reflect on their mistakes and generate better responses. It can be viewed as a process of in-context optimization to produce better responses. Previous works demonstrate that self-reflection can significantly improve LLM response quality \citep{renze2024self,shinn2024reflexion,madaan2024self}. LEPA utilizes self-reflection to optimize plans and solutions in the data generation phase, and acquires data of higher quality.

\section{Conclusion}
This paper presents the fundamental problem of what data should be generated in self-training algorithms. Inspired by cognitive science research and recent meta-learning advances, we propose a novel idea of learning abstract meta-knowledge in the form of anticipatory problem-solving plans.
Based on this idea, we propose a novel self-training algorithm, LEPA, which automatically generates and learns the anticipatory plans. Experiment results on several challenging reasoning benchmarks demonstrate the effectiveness of LEPA. An interesting future direction is to incorporate LEPA with more advanced model optimization methods such as RL. It is also worth exploring how well can LEPA perform on larger and more advanced LLMs, and how to scale LEPA to utilize more inference compute. Furthermore, as LLMs may solve simple problems without planning, an important future direction is to automatically identify complex problems that require planning from simple problems that can be easily solved without planning. This identification can avoid wasting compute resources and help the LLM solve problems more efficiently.

\section*{Acknowledgement}
This work is supported by National Natural Science Foundation of China (62176135), the National Key R\&D Program of China (2022ZD0161700), Shanghai Qi Zhi Institute Innovation Program SQZ202306 and the Tsinghua University Dushi Program.

\section*{Ethics Statement}
Concerns about safety and reliability are key points of discussion in the LLM community. The use of anticipatory plans in LLMs is a step towards making the models' actions more understandable and transparent to people. Yet, LEPA cannot guarantee that every solution will strictly match the plans it creates, which means further work is needed to solidify the trustworthiness of LLMs.

\bibliography{iclr2025_conference}
\bibliographystyle{iclr2025_conference}
\newpage
\appendix
\section{Detailed Prompts and Hyper-parameters}
\label{detail}
\begin{figure}[h]
    \centering
    \begin{subfigure}[b]{\textwidth}
        \includegraphics[width=\textwidth]{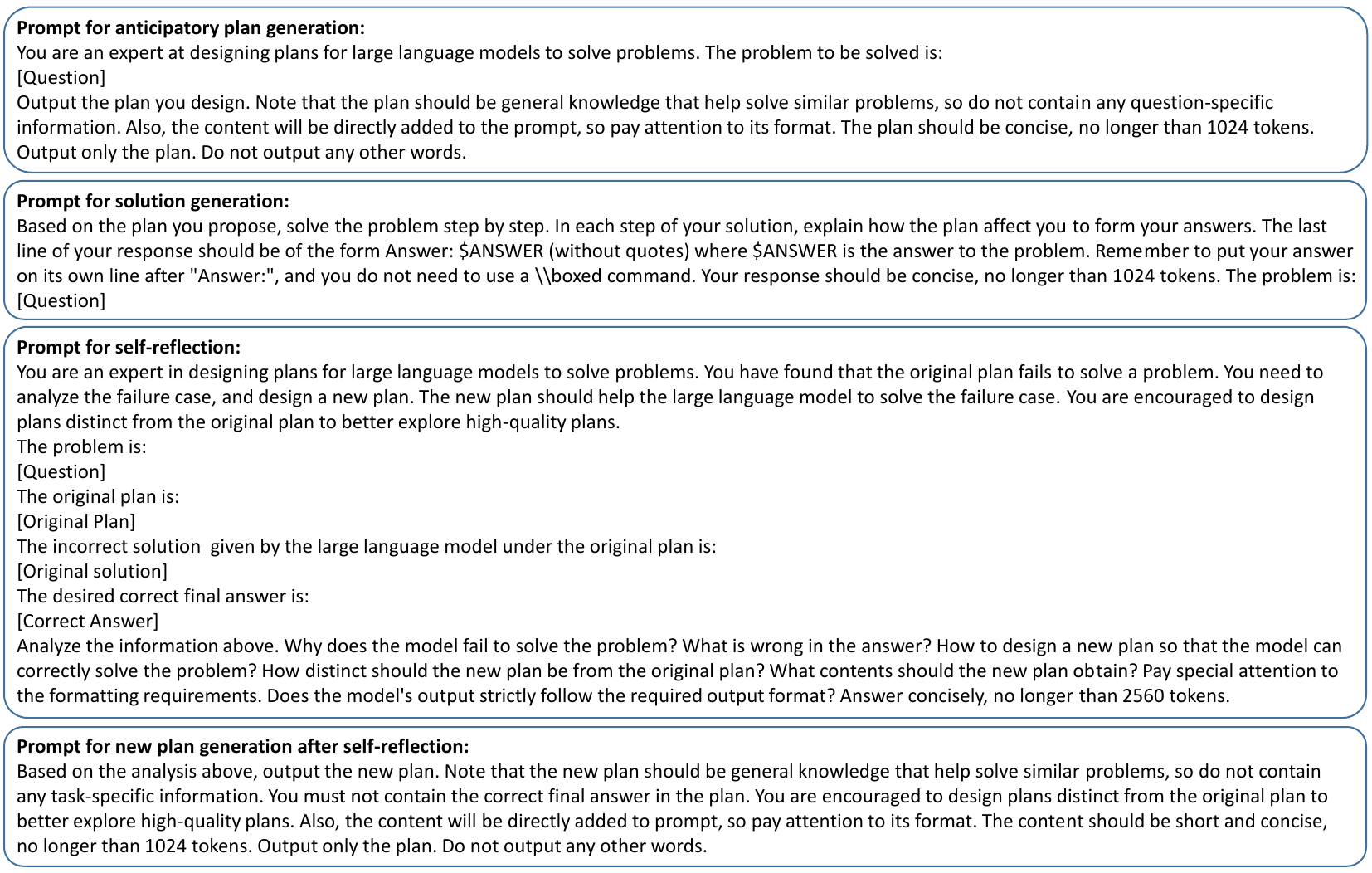}
        \caption{LEPA prompt.}
        \label{fig:p1}
    \end{subfigure}
    % \hfill % 可选，用于添加一些水平空间
    \begin{subfigure}[b]{\textwidth}
        \includegraphics[width=\textwidth]{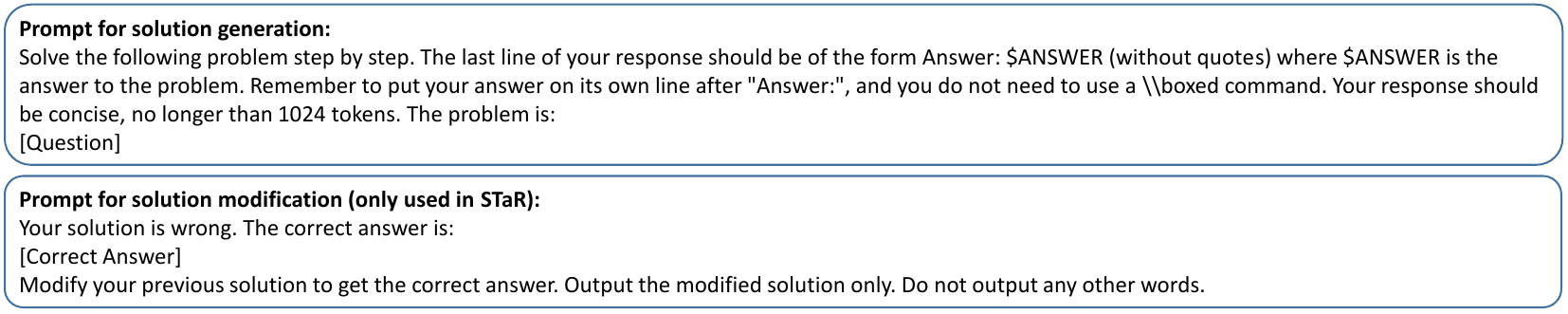}
        \caption{Prompt used by baseline methods.}
        \label{fig:p2}
    \end{subfigure}
    
    \caption{Detailed prompts used by (a) LEPA and (b) baseline algorithms.}
    \label{fig:prompt}
\end{figure}

% \subsection{Detailed Prompts}
This section demonstrates the detailed prompts and the hyper-parameters used by LEPA and baseline algorithms.
Figure \ref{fig:prompt} presents the prompts used by LEPA and baseline algorithms. 

As for hyper-parameters, for a fair comparison, we ensure that all algorithms have the same number of trials (5) in the data generation phase. LEPA is allowed to have maximally 4 self-reflection processes for each problem. For $ReST$ and $ReST^{EM}$, 5 solutions are sampled for each question. For STaR, it has maximally 4 opportunities to modify the previous incorrect answer. All algorithms fine-tunes the LLM for one epoch in each model optimization phase. For the data generation phase of all algorithms, we use a temperature of 0.5 for sampling. We use a temperature of 0.0005 for all test results. We use 3e-7 as the learning rate for all learning algorithms.

% \section{Case Study}
% \label{app2}
% The complete output of the case study is demonstrated in Figure \ref{}.

% \begin{figure}[t]
% \begin{center}
% %\framebox[4.0in]{$\;$}
% \includegraphics[width=\textwidth]{figures/case3.pdf}
% \end{center}
% \caption{A case study demonstrating how LEPA optimizes the anticipatory plans and the solutions with self-reflection. The initial plan is too broad and lacks detail, and fails to provide enough guidance to generate correct answers. The self-reflection process successfully analyses what is wrong, and generates a new, high-quality plan that provides more guidance while maintaining generalizability. With the new plan after self-reflection, the model successfully generates correct solutions.}
% \label{fig:case}
% \end{figure}

\section{Ablation Details}
\label{variant}
This section presents the details of the variants discussed in the ``Different ways of utilizing inference compute'' part of Section \ref{32}. 

For the second variant, we first sample correct and incorrect solutions for each problem with rejection sampling. Then we synthesize training data by first adding a sentence of ``Oops, I made a mistake. The correct solution is: '' to the end of incorrect solutions. Then we append a correct solution to the end of this sentence.

For the third variant, we explicitly instruct the LLM to output solutions that are approximately 2,000 words long. We observe that the LLM generates verbose responses that obscure the important steps in solving the problem.
\section{Additional Ablation Studies}
\label{aas}
\paragraph{OOD performance.} We evaluate OOD performance by training on Hendrycks MATH and testing on the Math category of MMLU-Pro \citep{wang2024mmlu}. As shown in Table \ref{tab:performance_comparison}, LEPA consistently outperforms baseline algorithms in this OOD setting.

\begin{table}[h]
\centering
\begin{tabular}{c|cc|ccc|c}
\hline
& {CoT} & {Plan+CoT} & {$ReST$} & {$ReST^{EM}$} & {STaR} & {LEPA} \\ \hline
Performance & 30.4\% & 33.9\% & 35.1\% & 35.3\% & 35.8\% & \textbf{38.9\%} \\ \hline
\end{tabular}
\caption{Performance of different algorithms training on Hendrycks MATH and testing on the Math category of MMLU-Pro.  "CoT" and "Plan+CoT" refer to the initial LLM's performance with a zero-shot CoT prompt and the LEPA prompt, respectively. LEPA achieves better generalization than baseline algorithms.}
\label{tab:performance_comparison}
\end{table}

\paragraph{Other LLMs.} We additionally evaluate algorithm performance on Llama 3.1 8B Instruct. As shown in Table 
\ref{tab:llama31}, on the Hendrycks MATH dataset, the LEPA prompt can slightly improve over the zero-shot CoT prompt on the initial LLM. As for self-training, LEPA significantly outperforms the baseline algorithm. These empirical results are consistent with our main results presented in Section \ref{31}.

\paragraph{Additional Benchmarks.} We additionally evaluate on CSQA \citep{saha2018complex} and MMLU \citep{hendrycks2020measuring}, and results are shown in Table 
\ref{tab:benchmark_comparison}. LEPA consistently outperforms baseline algorithms on these benchmarks.

\begin{table}[h]
\centering
\begin{tabular}{c|cc|ccc|c}
\hline
{Algorithm} & {CoT} & {Plan+CoT} & {$ReST$} & {$ReST^{EM}$} & {STaR} & {LEPA} \\ \hline
Performance & 37.2\% & 38.4\% & 45.3\% & 46.9\% & 45.0\% & \textbf{49.6\%} \\ \hline
\end{tabular}
\caption{Algorithm performance on Hendrycks MATH, with Llama 3.1 8B Instruct as the initial LLM. "CoT" and "Plan+CoT" refer to the initial LLM's performance with a zero-shot CoT prompt and the LEPA prompt, respectively}
\label{tab:llama31}
\end{table}

\begin{table}[h]
\centering
\begin{tabular}{c|cc|ccc|c}
\hline
 & {CoT} & {Plan+CoT} & {ReST} & {ReST$^{EM}$} & {STaR} & {LEPA} \\ \hline
CSQA & 67.1\% & 69.3\% & 73.2\% & 74.0\% & 74.1\% & \textbf{75.2\%} \\ \hline
MMLU & 61.9\% & 60.1\% & 64.3\% & 65.6\% & 65.8\% & \textbf{66.1\%} \\ \hline
\end{tabular}
\caption{Performance comparison of different methods on CSQA and MMLU benchmarks. LEPA achieves higher performance than baseline algorithms.}
\label{tab:benchmark_comparison}
\end{table}
\paragraph{Evaluation with Simple-Eval.} We re-evaluate Hendrycks MATH performance with Simple-Eval, and the results are demonstrated in Table \ref{tab:last}. With the new evaluation, LEPA still outperforms baseline algorithms.
\begin{table}[h]
\centering
\begin{tabular}{c|cc|ccc|c}
\hline
 & {CoT} & {Plan+CoT} & {$ReST$} & {$ReST^{EM}$} & {STaR} & {LEPA} \\ \hline
 & 26.1\% & 28.5\% &  31.2\% & 31.4\% & 29.2\% & \textbf{33.7\%} \\ \hline
\end{tabular}
\caption{Hendrycks MATH performance evaluated with Simple-Eval. With the new evaluation, LEPA still outperforms baseline algorithms. }
\label{tab:last}
\end{table}

\end{document}

%% file: math_commands.tex
\def\eqref#1{equation~\ref{#1}}
\def\1{\bm{1}}

\DeclareMathAlphabet{\mathsfit}{\encodingdefault}{\sfdefault}{m}{sl}
\SetMathAlphabet{\mathsfit}{bold}{\encodingdefault}{\sfdefault}{bx}{n}

%% file: iclr2025_conference.bbl
\begin{thebibliography}{48}
\providecommand{\natexlab}[1]{#1}
\providecommand{\url}[1]{\texttt{#1}}
\expandafter\ifx\csname urlstyle\endcsname\relax
  \providecommand{\doi}[1]{doi: #1}\else
  \providecommand{\doi}{doi: \begingroup \urlstyle{rm}\Url}\fi

\bibitem[Achiam et~al.(2023)Achiam, Adler, Agarwal, Ahmad, Akkaya, Aleman, Almeida, Altenschmidt, Altman, Anadkat, et~al.]{achiam2023gpt}
Josh Achiam, Steven Adler, Sandhini Agarwal, Lama Ahmad, Ilge Akkaya, Florencia~Leoni Aleman, Diogo Almeida, Janko Altenschmidt, Sam Altman, Shyamal Anadkat, et~al.
\newblock Gpt-4 technical report.
\newblock \emph{arXiv preprint arXiv:2303.08774}, 2023.

\bibitem[Bansal et~al.(2024)Bansal, Hosseini, Agarwal, Tran, and Kazemi]{bansal2024smaller}
Hritik Bansal, Arian Hosseini, Rishabh Agarwal, Vinh~Q Tran, and Mehran Kazemi.
\newblock Smaller, weaker, yet better: Training llm reasoners via compute-optimal sampling.
\newblock \emph{arXiv preprint arXiv:2408.16737}, 2024.

\bibitem[Bisk et~al.(2020)Bisk, Zellers, Gao, Choi, et~al.]{bisk2020piqa}
Yonatan Bisk, Rowan Zellers, Jianfeng Gao, Yejin Choi, et~al.
\newblock Piqa: Reasoning about physical commonsense in natural language.
\newblock In \emph{Proceedings of the AAAI conference on artificial intelligence}, volume~34, pp.\  7432--7439, 2020.

\bibitem[Cheng et~al.(2024)Cheng, Hu, Xu, Zhang, Dai, Han, and Du]{cheng2024self}
Pengyu Cheng, Tianhao Hu, Han Xu, Zhisong Zhang, Yong Dai, Lei Han, and Nan Du.
\newblock Self-playing adversarial language game enhances llm reasoning.
\newblock \emph{arXiv preprint arXiv:2404.10642}, 2024.

\bibitem[Clark et~al.(2019)Clark, Lee, Chang, Kwiatkowski, Collins, and Toutanova]{clark2019boolq}
Christopher Clark, Kenton Lee, Ming-Wei Chang, Tom Kwiatkowski, Michael Collins, and Kristina Toutanova.
\newblock Boolq: Exploring the surprising difficulty of natural yes/no questions.
\newblock \emph{arXiv preprint arXiv:1905.10044}, 2019.

\bibitem[Dou et~al.(2024)Dou, Yang, Wu, Chang, and Peng]{dou2024reflection}
Zi-Yi Dou, Cheng-Fu Yang, Xueqing Wu, Kai-Wei Chang, and Nanyun Peng.
\newblock Reflection-reinforced self-training for language agents.
\newblock \emph{arXiv preprint arXiv:2406.01495}, 2024.

\bibitem[Dubey et~al.(2024)Dubey, Jauhri, Pandey, Kadian, Al-Dahle, Letman, Mathur, Schelten, Yang, Fan, et~al.]{dubey2024llama}
Abhimanyu Dubey, Abhinav Jauhri, Abhinav Pandey, Abhishek Kadian, Ahmad Al-Dahle, Aiesha Letman, Akhil Mathur, Alan Schelten, Amy Yang, Angela Fan, et~al.
\newblock The llama 3 herd of models.
\newblock \emph{arXiv preprint arXiv:2407.21783}, 2024.

\bibitem[Finn et~al.(2017)Finn, Abbeel, and Levine]{finn2017model}
Chelsea Finn, Pieter Abbeel, and Sergey Levine.
\newblock Model-agnostic meta-learning for fast adaptation of deep networks.
\newblock In \emph{International conference on machine learning}, pp.\  1126--1135. PMLR, 2017.

\bibitem[Fu et~al.(2024)Fu, Huangfu, Yan, Ng, and Qiu]{fu2024hint}
Jinlan Fu, Shenzhen Huangfu, Hang Yan, See-Kiong Ng, and Xipeng Qiu.
\newblock Hint-before-solving prompting: Guiding llms to effectively utilize encoded knowledge.
\newblock \emph{arXiv preprint arXiv:2402.14310}, 2024.

\bibitem[Goyal et~al.(2023)Goyal, Ji, Rawat, Menon, Kumar, and Nagarajan]{goyal2023think}
Sachin Goyal, Ziwei Ji, Ankit~Singh Rawat, Aditya~Krishna Menon, Sanjiv Kumar, and Vaishnavh Nagarajan.
\newblock Think before you speak: Training language models with pause tokens.
\newblock \emph{arXiv preprint arXiv:2310.02226}, 2023.

\bibitem[Gulcehre et~al.(2023)Gulcehre, Paine, Srinivasan, Konyushkova, Weerts, Sharma, Siddhant, Ahern, Wang, Gu, et~al.]{gulcehre2023reinforced}
Caglar Gulcehre, Tom~Le Paine, Srivatsan Srinivasan, Ksenia Konyushkova, Lotte Weerts, Abhishek Sharma, Aditya Siddhant, Alex Ahern, Miaosen Wang, Chenjie Gu, et~al.
\newblock Reinforced self-training (rest) for language modeling.
\newblock \emph{arXiv preprint arXiv:2308.08998}, 2023.

\bibitem[Hendrycks et~al.(2020)Hendrycks, Burns, Basart, Zou, Mazeika, Song, and Steinhardt]{hendrycks2020measuring}
Dan Hendrycks, Collin Burns, Steven Basart, Andy Zou, Mantas Mazeika, Dawn Song, and Jacob Steinhardt.
\newblock Measuring massive multitask language understanding.
\newblock \emph{arXiv preprint arXiv:2009.03300}, 2020.

\bibitem[Hendrycks et~al.(2021)Hendrycks, Burns, Kadavath, Arora, Basart, Tang, Song, and Steinhardt]{hendrycks2021measuring}
Dan Hendrycks, Collin Burns, Saurav Kadavath, Akul Arora, Steven Basart, Eric Tang, Dawn Song, and Jacob Steinhardt.
\newblock Measuring mathematical problem solving with the math dataset.
\newblock \emph{arXiv preprint arXiv:2103.03874}, 2021.

\bibitem[Hoffman et~al.(2024)Hoffman, Phan, Dohan, Douglas, Le, Parisi, Sountsov, Sutton, Vikram, and A~Saurous]{hoffman2024training}
Matthew~Douglas Hoffman, Du~Phan, David Dohan, Sholto Douglas, Tuan~Anh Le, Aaron Parisi, Pavel Sountsov, Charles Sutton, Sharad Vikram, and Rif A~Saurous.
\newblock Training chain-of-thought via latent-variable inference.
\newblock \emph{Advances in Neural Information Processing Systems}, 36, 2024.

\bibitem[Hu et~al.(2024)Hu, Shang, Xu, He, and Zhang]{hu2024can}
Haichuan Hu, Ye~Shang, Guolin Xu, Congqing He, and Quanjun Zhang.
\newblock Can gpt-o1 kill all bugs?
\newblock \emph{arXiv preprint arXiv:2409.10033}, 2024.

\bibitem[Huang et~al.(2022)Huang, Gu, Hou, Wu, Wang, Yu, and Han]{huang2022large}
Jiaxin Huang, Shixiang~Shane Gu, Le~Hou, Yuexin Wu, Xuezhi Wang, Hongkun Yu, and Jiawei Han.
\newblock Large language models can self-improve.
\newblock \emph{arXiv preprint arXiv:2210.11610}, 2022.

\bibitem[Kojima et~al.(2022)Kojima, Gu, Reid, Matsuo, and Iwasawa]{kojima2022large}
Takeshi Kojima, Shixiang~Shane Gu, Machel Reid, Yutaka Matsuo, and Yusuke Iwasawa.
\newblock Large language models are zero-shot reasoners.
\newblock \emph{Advances in neural information processing systems}, 35:\penalty0 22199--22213, 2022.

\bibitem[Madaan et~al.(2024)Madaan, Tandon, Gupta, Hallinan, Gao, Wiegreffe, Alon, Dziri, Prabhumoye, Yang, et~al.]{madaan2024self}
Aman Madaan, Niket Tandon, Prakhar Gupta, Skyler Hallinan, Luyu Gao, Sarah Wiegreffe, Uri Alon, Nouha Dziri, Shrimai Prabhumoye, Yiming Yang, et~al.
\newblock Self-refine: Iterative refinement with self-feedback.
\newblock \emph{Advances in Neural Information Processing Systems}, 36, 2024.

\bibitem[Rad{\"u}ntz(2020)]{raduntz2020effect}
Thea Rad{\"u}ntz.
\newblock The effect of planning, strategy learning, and working memory capacity on mental workload.
\newblock \emph{Scientific reports}, 10\penalty0 (1):\penalty0 7096, 2020.

\bibitem[Rafailov et~al.(2024)Rafailov, Sharma, Mitchell, Manning, Ermon, and Finn]{rafailov2024direct}
Rafael Rafailov, Archit Sharma, Eric Mitchell, Christopher~D Manning, Stefano Ermon, and Chelsea Finn.
\newblock Direct preference optimization: Your language model is secretly a reward model.
\newblock \emph{Advances in Neural Information Processing Systems}, 36, 2024.

\bibitem[Rakelly et~al.(2019)Rakelly, Zhou, Finn, Levine, and Quillen]{rakelly2019efficient}
Kate Rakelly, Aurick Zhou, Chelsea Finn, Sergey Levine, and Deirdre Quillen.
\newblock Efficient off-policy meta-reinforcement learning via probabilistic context variables.
\newblock In \emph{International conference on machine learning}, pp.\  5331--5340. PMLR, 2019.

\bibitem[Renze \& Guven(2024)Renze and Guven]{renze2024self}
Matthew Renze and Erhan Guven.
\newblock Self-reflection in llm agents: Effects on problem-solving performance.
\newblock \emph{arXiv preprint arXiv:2405.06682}, 2024.

\bibitem[Ross(2009)]{ross2009psychology}
Brian~H Ross.
\newblock The psychology of learning and motivation: Advances in research and theory.
\newblock 2009.

\bibitem[Saha et~al.(2018)Saha, Pahuja, Khapra, Sankaranarayanan, and Chandar]{saha2018complex}
Amrita Saha, Vardaan Pahuja, Mitesh Khapra, Karthik Sankaranarayanan, and Sarath Chandar.
\newblock Complex sequential question answering: Towards learning to converse over linked question answer pairs with a knowledge graph.
\newblock In \emph{Proceedings of the AAAI conference on artificial intelligence}, volume~32, 2018.

\bibitem[Schulman et~al.(2017)Schulman, Wolski, Dhariwal, Radford, and Klimov]{schulman2017proximal}
John Schulman, Filip Wolski, Prafulla Dhariwal, Alec Radford, and Oleg Klimov.
\newblock Proximal policy optimization algorithms.
\newblock \emph{arXiv preprint arXiv:1707.06347}, 2017.

\bibitem[Shahriar et~al.(2024)Shahriar, Lund, Mannuru, Arshad, Hayawi, Bevara, Mannuru, and Batool]{shahriar2024putting}
Sakib Shahriar, Brady~D Lund, Nishith~Reddy Mannuru, Muhammad~Arbab Arshad, Kadhim Hayawi, Ravi Varma~Kumar Bevara, Aashrith Mannuru, and Laiba Batool.
\newblock Putting gpt-4o to the sword: A comprehensive evaluation of language, vision, speech, and multimodal proficiency.
\newblock \emph{Applied Sciences}, 14\penalty0 (17):\penalty0 7782, 2024.

\bibitem[Shinn et~al.(2024)Shinn, Cassano, Gopinath, Narasimhan, and Yao]{shinn2024reflexion}
Noah Shinn, Federico Cassano, Ashwin Gopinath, Karthik Narasimhan, and Shunyu Yao.
\newblock Reflexion: Language agents with verbal reinforcement learning.
\newblock \emph{Advances in Neural Information Processing Systems}, 36, 2024.

\bibitem[Singh et~al.(2023)Singh, Co-Reyes, Agarwal, Anand, Patil, Liu, Harrison, Lee, Xu, Parisi, et~al.]{singh2023beyond}
Avi Singh, John~D Co-Reyes, Rishabh Agarwal, Ankesh Anand, Piyush Patil, Peter~J Liu, James Harrison, Jaehoon Lee, Kelvin Xu, Aaron Parisi, et~al.
\newblock Beyond human data: Scaling self-training for problem-solving with language models.
\newblock \emph{arXiv preprint arXiv:2312.06585}, 2023.

\bibitem[Snell et~al.(2024)Snell, Lee, Xu, and Kumar]{snell2024scaling}
Charlie Snell, Jaehoon Lee, Kelvin Xu, and Aviral Kumar.
\newblock Scaling llm test-time compute optimally can be more effective than scaling model parameters.
\newblock \emph{arXiv preprint arXiv:2408.03314}, 2024.

\bibitem[Sung et~al.(2017)Sung, Zhang, Xiang, Hospedales, and Yang]{sung2017learning}
Flood Sung, Li~Zhang, Tao Xiang, Timothy Hospedales, and Yongxin Yang.
\newblock Learning to learn: Meta-critic networks for sample efficient learning.
\newblock \emph{arXiv preprint arXiv:1706.09529}, 2017.

\bibitem[Wang et~al.(2023{\natexlab{a}})Wang, Zhang, Jiang, Zhang, Wang, and Zhang]{wang2023offline}
Jianhao Wang, Jin Zhang, Haozhe Jiang, Junyu Zhang, Liwei Wang, and Chongjie Zhang.
\newblock Offline meta reinforcement learning with in-distribution online adaptation.
\newblock In \emph{International Conference on Machine Learning}, pp.\  36626--36669. PMLR, 2023{\natexlab{a}}.

\bibitem[Wang et~al.(2023{\natexlab{b}})Wang, Xu, Lan, Hu, Lan, Lee, and Lim]{wang2023plan}
Lei Wang, Wanyu Xu, Yihuai Lan, Zhiqiang Hu, Yunshi Lan, Roy Ka-Wei Lee, and Ee-Peng Lim.
\newblock Plan-and-solve prompting: Improving zero-shot chain-of-thought reasoning by large language models.
\newblock \emph{arXiv preprint arXiv:2305.04091}, 2023{\natexlab{b}}.

\bibitem[Wang \& Chiew(2010)Wang and Chiew]{wang2010cognitive}
Yingxu Wang and Vincent Chiew.
\newblock On the cognitive process of human problem solving.
\newblock \emph{Cognitive systems research}, 11\penalty0 (1):\penalty0 81--92, 2010.

\bibitem[Wang et~al.(2024{\natexlab{a}})Wang, Ma, Zhang, Ni, Chandra, Guo, Ren, Arulraj, He, Jiang, et~al.]{wang2024mmlu}
Yubo Wang, Xueguang Ma, Ge~Zhang, Yuansheng Ni, Abhranil Chandra, Shiguang Guo, Weiming Ren, Aaran Arulraj, Xuan He, Ziyan Jiang, et~al.
\newblock Mmlu-pro: A more robust and challenging multi-task language understanding benchmark.
\newblock \emph{arXiv preprint arXiv:2406.01574}, 2024{\natexlab{a}}.

\bibitem[Wang et~al.(2024{\natexlab{b}})Wang, Bi, Pentyala, Ramnath, Chaudhuri, Mehrotra, Mao, Asur, et~al.]{wang2024comprehensive}
Zhichao Wang, Bin Bi, Shiva~Kumar Pentyala, Kiran Ramnath, Sougata Chaudhuri, Shubham Mehrotra, Xiang-Bo Mao, Sitaram Asur, et~al.
\newblock A comprehensive survey of llm alignment techniques: Rlhf, rlaif, ppo, dpo and more.
\newblock \emph{arXiv preprint arXiv:2407.16216}, 2024{\natexlab{b}}.

\bibitem[Wang et~al.(2024{\natexlab{c}})Wang, Cai, Chen, Liu, Ma, and Liang]{wang2024describe}
Zihao Wang, Shaofei Cai, Guanzhou Chen, Anji Liu, Xiaojian~Shawn Ma, and Yitao Liang.
\newblock Describe, explain, plan and select: interactive planning with llms enables open-world multi-task agents.
\newblock \emph{Advances in Neural Information Processing Systems}, 36, 2024{\natexlab{c}}.

\bibitem[Xiao et~al.(2023)Xiao, Lin, Seznec, Wu, Demouth, and Han]{xiao2023smoothquant}
Guangxuan Xiao, Ji~Lin, Mickael Seznec, Hao Wu, Julien Demouth, and Song Han.
\newblock Smoothquant: Accurate and efficient post-training quantization for large language models.
\newblock In \emph{International Conference on Machine Learning}, pp.\  38087--38099. PMLR, 2023.

\bibitem[Yang et~al.(2024)Yang, Yang, Hui, Zheng, Yu, Zhou, Li, Li, Liu, Huang, et~al.]{yang2024qwen2}
An~Yang, Baosong Yang, Binyuan Hui, Bo~Zheng, Bowen Yu, Chang Zhou, Chengpeng Li, Chengyuan Li, Dayiheng Liu, Fei Huang, et~al.
\newblock Qwen2 technical report.
\newblock \emph{arXiv preprint arXiv:2407.10671}, 2024.

\bibitem[Yao et~al.(2022)Yao, Zhao, Yu, Du, Shafran, Narasimhan, and Cao]{yao2022react}
Shunyu Yao, Jeffrey Zhao, Dian Yu, Nan Du, Izhak Shafran, Karthik Narasimhan, and Yuan Cao.
\newblock React: Synergizing reasoning and acting in language models.
\newblock \emph{arXiv preprint arXiv:2210.03629}, 2022.

\bibitem[Yin et~al.(2023)Yin, Brahman, Ravichander, Chandu, Chang, Choi, and Lin]{yin2023lumos}
Da~Yin, Faeze Brahman, Abhilasha Ravichander, Khyathi Chandu, Kai-Wei Chang, Yejin Choi, and Bill~Yuchen Lin.
\newblock Lumos: Learning agents with unified data, modular design, and open-source llms.
\newblock In \emph{ICLR 2024 Workshop on Large Language Model (LLM) Agents}, 2023.

\bibitem[Yuan et~al.(2023)Yuan, Yuan, Li, Dong, Lu, Tan, Zhou, and Zhou]{yuan2023scaling}
Zheng Yuan, Hongyi Yuan, Chengpeng Li, Guanting Dong, Keming Lu, Chuanqi Tan, Chang Zhou, and Jingren Zhou.
\newblock Scaling relationship on learning mathematical reasoning with large language models.
\newblock \emph{arXiv preprint arXiv:2308.01825}, 2023.

\bibitem[Zelikman et~al.(2022)Zelikman, Wu, Mu, and Goodman]{zelikman2022star}
Eric Zelikman, Yuhuai Wu, Jesse Mu, and Noah Goodman.
\newblock Star: Bootstrapping reasoning with reasoning.
\newblock \emph{Advances in Neural Information Processing Systems}, 35:\penalty0 15476--15488, 2022.

\bibitem[Zelikman et~al.(2024)Zelikman, Harik, Shao, Jayasiri, Haber, and Goodman]{zelikman2024quiet}
Eric Zelikman, Georges Harik, Yijia Shao, Varuna Jayasiri, Nick Haber, and Noah~D Goodman.
\newblock Quiet-star: Language models can teach themselves to think before speaking.
\newblock \emph{arXiv preprint arXiv:2403.09629}, 2024.

\bibitem[Zellers et~al.(2019)Zellers, Holtzman, Bisk, Farhadi, and Choi]{zellers2019hellaswag}
Rowan Zellers, Ari Holtzman, Yonatan Bisk, Ali Farhadi, and Yejin Choi.
\newblock Hellaswag: Can a machine really finish your sentence?
\newblock \emph{arXiv preprint arXiv:1905.07830}, 2019.

\bibitem[Zhang et~al.(2021{\natexlab{a}})Zhang, Wang, Hu, Chen, Chen, Fan, and Zhang]{zhang2021metacure}
Jin Zhang, Jianhao Wang, Hao Hu, Tong Chen, Yingfeng Chen, Changjie Fan, and Chongjie Zhang.
\newblock Metacure: Meta reinforcement learning with empowerment-driven exploration.
\newblock In \emph{International Conference on Machine Learning}, pp.\  12600--12610. PMLR, 2021{\natexlab{a}}.

\bibitem[Zhang et~al.(2021{\natexlab{b}})Zhang, Kim, O'Donoghue, and Boyd]{zhang2021sample}
Junzi Zhang, Jongho Kim, Brendan O'Donoghue, and Stephen Boyd.
\newblock Sample efficient reinforcement learning with reinforce.
\newblock In \emph{Proceedings of the AAAI conference on artificial intelligence}, volume~35, pp.\  10887--10895, 2021{\natexlab{b}}.

\bibitem[Zhao et~al.(2023)Zhao, Zhou, Li, Tang, Wang, Hou, Min, Zhang, Zhang, Dong, et~al.]{zhao2023survey}
Wayne~Xin Zhao, Kun Zhou, Junyi Li, Tianyi Tang, Xiaolei Wang, Yupeng Hou, Yingqian Min, Beichen Zhang, Junjie Zhang, Zican Dong, et~al.
\newblock A survey of large language models.
\newblock \emph{arXiv preprint arXiv:2303.18223}, 2023.

\bibitem[Zheng et~al.(2023)Zheng, Liu, Xie, Li, and Li]{zheng2023progressive}
Chuanyang Zheng, Zhengying Liu, Enze Xie, Zhenguo Li, and Yu~Li.
\newblock Progressive-hint prompting improves reasoning in large language models.
\newblock \emph{arXiv preprint arXiv:2304.09797}, 2023.

\end{thebibliography}
